%% file: main.tex
\pdfoutput=1

\documentclass[11pt]{article}

\newcommand{\paperreview}{review}
\newcommand{\paperfinal}{final}

\let\paperversion\paperfinal
\usepackage[\paperversion]{styles/acl}

\usepackage{times}
\usepackage{latexsym}

\usepackage[T1]{fontenc}

\usepackage[utf8]{inputenc}

\usepackage{microtype}

\usepackage{inconsolata}

\usepackage{graphicx}

\usepackage{algorithm}
\usepackage{algpseudocode}
\usepackage{amsmath}
\usepackage{amssymb}
\usepackage{listingsutf8}
\usepackage{booktabs}
\usepackage{multirow}
\usepackage{array}
\usepackage{longtable}
\usepackage{xcolor}
\usepackage{authblk}
\usepackage{hyperref}

\newcounter{snippetpython}
\newcounter{unflavoredsnippet}

\makeatletter

\lstnewenvironment{unflavoredsnippet}[1][]{
    
    \let\c@lstlisting=\c@unflavoredsnippet
    
    \lstset{
        inputencoding=utf8/latin1,
        basicstyle=\small\ttfamily,
        columns=flexible,
        breaklines=true,
        breakatwhitespace=true,
        breakindent=0em,
        aboveskip=-0.9em,
        belowskip=-0.9em,
        literate=
            {«}{\guillemotleft{}}1
            {»}{\guillemotright{}}1
            {ú}{\'{u}}1
            {ü}{{\"u}}1,
        moredelim=**[is][\color{ModelCompletion}]{\#}{\#},
        #1
    }
}{}

\lstnewenvironment{snippetpython}[1][]{
    
    \let\c@lstlisting=\c@snippetpython
    
    \lstset{
        language=Python,
        basicstyle=\ttfamily\footnotesize,
        captionpos=b,
        xleftmargin=2pt,
        xrightmargin=2pt,
        numbersep=2pt,
        numbers=left,
        numberstyle=\tiny\color{gray},
        showstringspaces=false,
        breaklines=true,
        keywordstyle=\color{DarkCyan},
        stringstyle=\color{OrangeWeb},
        commentstyle=\color{Avocado},
        backgroundcolor=\color{gray!5},
        morekeywords={dspy, self},
        #1
    }
}{}

\makeatother

\definecolor{Avocado}{HTML}{688E26}
\definecolor{CaribbeanCurrent}{HTML}{187077}
\definecolor{DarkCyan}{HTML}{0E9594}
\definecolor{OrangeWeb}{HTML}{FFA400}
\definecolor{PigmentGreen}{HTML}{08A045}

\colorlet{PrimaryColor}{CaribbeanCurrent}
\colorlet{SecondaryColor}{OrangeWeb}
\colorlet{ModelCompletion}{PigmentGreen}

\hypersetup{
    allcolors={PrimaryColor},
    linkcolor={SecondaryColor},
    citecolor={PrimaryColor!80},
}

\newcommand{\bettertogether}{\textcolor{PrimaryColor}{\textbf{\texttt{BetterTogether}}}}

\newcommand{\LanguageProgram}[1][\empty]{
    \Phi_{\langle \Theta, \Pi \rangle \ifx#1\empty \else \gets #1\fi}
}
\newcommand{\LanguageProgramWeights}{\Phi_{\Theta}}
\newcommand{\LanguageProgramPrompts}{\Phi_{\Pi}}

\DeclareMathOperator*{\argmax}{arg\,max}

\newcommand{\hotpotqa}{\texttt{HotPotQA}}
\newcommand{\gsmk}{\texttt{GSM8K}}
\newcommand{\iris}{\texttt{Iris}}

\newcommand{\colberttwo}{\texttt{ColBERTv2}}
\newcommand{\mistral}{\texttt{mistralai/Mistral-7b-Instruct-v0.2}}
\newcommand{\mistralshort}{\texttt{mistral-7b-instruct-v0.2}}
\newcommand{\mistralshorter}{\texttt{mistral-7b}}
\newcommand{\llamatwo}{\texttt{meta-llama/Llama-2-7b-chat-hf}}
\newcommand{\llamatwoshort}{\texttt{llama-2-7b-chat}}
\newcommand{\llamatwoshorter}{\texttt{llama-2-7b}}
\newcommand{\llamathree}{\texttt{meta-llama/Meta-Llama-3-8B-Instruct}}
\newcommand{\llamathreeshort}{\texttt{llama-3-8b-instruct}}
\newcommand{\llamathreeshorter}{\texttt{llama-3-8b}}

\newcommand{\apachetwolicense}{\texttt{Apache License 2.0}}
\newcommand{\mitlicense}{\texttt{MIT License}}
\newcommand{\ccbyfour}{\texttt{Creative Commons Attribution 4.0 International (CC BY 4.0)}}
\newcommand{\llamatwocommunitylicense}{\texttt{Meta Llama 2 Community License} at \url{https://ai.meta.com/llama/license/}}
\newcommand{\llamathreecommunitylicense}{\texttt{Meta Llama 3 Community License} at \url{https://llama.meta.com/llama3/license/}}

\newcommand{\gpuaonehundred}{\texttt{A100}}

\newcommand{\defaulttrain}{$1000$}
\newcommand{\defaultdev}{$500$}
\newcommand{\defaultprompttrain}{$100$}
\newcommand{\defaultpromptval}{$250$}
\newcommand{\hotpotqatest}{$1500$}
\newcommand{\gsmktest}{$1319$}
\newcommand{\iristotal}{$150$}
\newcommand{\iriscustomsplit}{$50$}
\newcommand{\iriscustomprompttrain}{$15$}
\newcommand{\iriscustompromptval}{$35$}

\newcommand{\finetunelr}{$1\mathrm{e}{-5}$}
\newcommand{\finetuneeffectivebatchsize}{$8$}
\newcommand{\finetuneprecision}{\texttt{bfloat16}}
\newcommand{\finetunenumepochs}{$5$}
\newcommand{\finetunelorar}{$32$}
\newcommand{\finetuneloraalpha}{$64$}
\newcommand{\finetuneloratarget}{\texttt{query} and \texttt{key} self-attention layers}

\newcommand{\promptmaxlabeledemos}{$3$}
\newcommand{\promptnumcandidateprograms}{$6$}

\newcommand{\samplingtemperature}{$0.1$}
\newcommand{\samplingtopk}{$0.97$}
\newcommand{\samplingmaxtokens}{$1024$}

\newcommand{\gainsweightsoverallupto}{60\%}
\newcommand{\gainspromptsoverallupto}{6\%}
\newcommand{\gainshotpotqaminmax}{5--78\%}
\newcommand{\gainsgsmkminmax}{2.5--10\%}
\newcommand{\gainsirisminmax}{3.5--88\%}
\newcommand{\resultsuccessrate}{7 out of the 9}

\title{
    Fine-Tuning and Prompt Optimization:
    \\ [0.5ex]
    Two Great Steps that Work \textcolor{PrimaryColor}{Better Together}
}

\author{\textbf{Dilara Soylu}}
\author{\textbf{Christopher Potts}}
\author{\textbf{Omar Khattab}}
\affil{Stanford University}
\affil{\texttt{\{soylu,cgpotts,okhattab\}@stanford.edu}}

\begin{document}

\maketitle

\begin{abstract}
    \input{sections/__abstract}

\end{abstract}

\input{sections/01_introduction}
\input{sections/02_methods}
\input{sections/03_experiments}
\input{sections/04_conclusion}
\input{sections/05_limitations}

\ifx\paperversion\paperreview
\else
  \input{sections/__acknowledgments}
\fi

\bibliography{bibliography/anthology, bibliography/custom}

\newpage
\appendix

\onecolumn
\renewcommand{\thetable}{\Alph{section}.\arabic{table}}
\numberwithin{table}{section}
\renewcommand{\thesnippetpython}{\Alph{section}.\arabic{snippetpython}}
\numberwithin{snippetpython}{section}
\renewcommand{\theunflavoredsnippet}{\Alph{section}.\arabic{unflavoredsnippet}}
\numberwithin{unflavoredsnippet}{section}

\section*{Appendices}
\input{appendices/01_programs}

\input{appendices/02_assets}
\input{appendices/03_implementation}
\input{appendices/04_extended_results}

\end{document}

%% file: sections/__abstract.tex
Natural Language Processing (NLP) systems are increasingly taking the form of sophisticated modular pipelines, e.g., Retrieval Augmented Generation (RAG), where each module may involve a distinct Language Model (LM) and an associated prompt template.
These compound systems often lack intermediate labels or gradient flow to optimize each module, making their end-to-end optimization challenging.
Here we seek strategies to optimize both the module-level LM weights and the associated prompt templates of such systems to maximize a downstream task metric.
We propose for the first time combining the weight and prompt optimization strategies to optimize a modular LM pipeline by alternating between the two to get the same LM to teach itself.
In experiments with multi-hop QA, mathematical reasoning, and feature-based classification using \mistralshorter, \llamatwoshorter, and \llamathreeshorter, these \bettertogether\ strategies optimizing the weights and prompts of a pipeline together outperform directly optimizing weights alone and prompts alone by up to \gainsweightsoverallupto\ and \gainspromptsoverallupto, respectively, on average across LMs and tasks.
\ifx\paperversion\paperreview
  Our \bettertogether\ optimizer is released in [link redacted for this version].
\else
  Our \bettertogether\ optimizer is released in DSPy at \url{http://dspy.ai}.
\fi

%% file: sections/01_introduction.tex
\section{Introduction}
\label{sec:introduction}

While the capabilities of language models (LMs) continue to grow, recent work has shown the potential of building more powerful Natural Language Processing (NLP) systems by \textit{composing} multiple skills of LMs into pipelines. Examples of this include systems for retrieval-augmented generation~\cite{guu2020realm,lewis2020retrieval,ma-etal-2023-query,jiang-etal-2023-active}, multi-hop reasoning~\cite{qi2021retrieve,khattab2021baleen}, information extraction~\cite{pourreza2023dinsql,doosterlinck2024incontext}, and other sophisticated pipelines~\cite{zelikman2022star,dohan2022language,khattab2022demonstrate,beurerkellner2023prompting,schlag2023large,viswanathan2023prompt2model}.

Such \textit{LM Programs} offer much more control for designing NLP systems, as they break down problems into modular, more manageable sub-tasks that can be assigned to LMs. If we could teach these LMs to accurately conduct their easier sub-tasks and to communicate effectively within multi-stage pipelines, this could greatly expand the scope of reliable NLP systems we can build. 

To this end, \citet{khattab2024dspy} introduced the DSPy framework for defining and \textit{optimizing} LM Programs.
In it, a program is defined as a function $\Phi$ that composes a set of stages, which we will refer to as \textit{language modules} $M = \langle M_{1}, \ldots, M_{|M|} \rangle$, into a pipeline.
Each language module $M_i$ specifies a fuzzy natural-language transformation (e.g., generating a summary of a supplied document) that needs to be learned.
To do so, each module learns a particular prompt (template) $\pi$ to make a call to a particular LM with weights $\theta$.
The optimization problem is then defined as maximizing the expected performance (per a downstream metric $\mu$) of the program $\Phi$ over a set of inputs by updating each module's $\pi$ and $\theta$.

Existing work~\cite{khattab2024dspy,opsahlong2024optimizing} has studied optimizing the discrete string prompt of each module and has considered simple approaches for fine-tuning each module's LM weights.
In this empirical study, we investigate \textit{updating each module's LM weights and prompt template together} to maximize a downstream metric on the final output of the program.
Doing this is challenging as $\Phi$ is not generally differentiable and its modules $M_i$ generally lack labeled outputs, while exhibiting sophisticated dependencies. Moreover, in realistic settings, the training set is usually very small and only a small number of LM calls are possible for training and inference.

To address this challenge, we propose to \textit{alternate between fine-tuning LM weights and optimizing prompt templates} and evaluate approximate optimization strategies in which we bootstrap training labels for all pipeline modules.
In experiments with multi-hop QA (\hotpotqa), mathematical reasoning (\gsmk), and feature-based classification (\iris), we show that these tandem strategies are highly effective across three different LMs, leading to \gainshotpotqaminmax\ gains for \hotpotqa, \gainsgsmkminmax\ gains for \gsmk, and \gainsirisminmax\ gains for \iris\ against prompts only and weights only strategies, averaged across \mistralshort, \llamatwoshort, and \llamathreeshort.

%% file: sections/02_methods.tex
\section{Problem Statement}
\label{sec:problem}

We are given an LM program $\Phi$, which operates like a blackbox function $\Phi : \mathcal{X} \to \mathcal{Y}$, in which $\mathcal{X}$ and $\mathcal{Y}$ are typically in natural language (e.g., questions and their program-generated answers, respectively).
For example, we may have a program $\Phi$ for answering complex questions with short factoid answers. In the course of its execution, $\Phi$ makes one or more calls to each of its $|M| \geq 1$ \textit{language modules}, $M = \langle M_{1}, \ldots, M_{|M|} \rangle$.

For example, the program may implement a \textit{multi-hop, retrieval-augmented} pipeline for question answering. This common pipeline~\cite{qi2021retrieve,khattab2021baleen,press2023measuring,khattab2022demonstrate} breaks down the input into sub-questions that are used to iteratively find relevant passages (e.g., from a corpus like Wikipedia) until the question can be faithfully answered. In general terms, each module $M_i : \mathcal{X}_{i} \to \mathcal{Y}_i$ is a \textit{declarative} LM invocation that defines, in inherently fuzzy natural-language terms, an input $\mathcal{X}_i$ domain (like a \textit{user-supplied question} and \textit{a set of retrieved passages}) and an output $\mathcal{Y}_{i}$ co-domain (like a \textit{search query to find additional relevant passages}).

We seek to implement each language module as some specific, well-tuned strategy for invoking an underlying language model $\mathbf{LM}$.
Concretely, we assume that a module $M_i$ will be fully implemented by specifying (1) the string prompt $\pi_i$ in which the module inputs $\mathcal{X}_i$ are plugged in to decode the module outputs $\mathcal{Y}_i$ and (2) the floating-point weights $\theta_i$ assigned to the parameters of $\mathbf{LM}$ in the course of this module.
We refer to the version of $\Phi$ in which the prompts and LM weights are assigned explicitly to $\Pi$ and $\Theta$, respectively, as~$\LanguageProgram$.

Given nothing but a small training set $X = \{ (x_{1}, m_{1}), \ldots, (x_{|X|}, m_{|X|)}) \}$ of inputs $x_{i} \in \mathcal{X}$ and optional \textit{metadata} like output labels or other \textit{hints} $m_{i} \in \mathcal{M}$ that can be used for determining the correctness of a given program run, and a metric $\mu : \mathcal{Y} \times \mathcal{M} \to \mathbb{R}$, our goal is to optimize $\Phi$, that is, configure its modules' prompts and LM weights to maximize the following objective.

\begin{equation*}
  \argmax_{\Theta, \Pi}
  \, \frac{1}{|X|} \sum_{(x, m)  \in X}
  \mu(\LanguageProgram(x), m)
  \label{eq:optimization_objective}
\end{equation*}

Researchers tuning LM pipelines are in effect seeking to achieve this objective.
It is also a very large subspace of the optimization problem in the DSPy framework\footnote{\url{http://dspy.ai}} for LM programs.
Unfortunately, this problem is intractable: the search space is large and we don't have gradients or intermediate output labels to optimize each module, so we seek approximate strategies for such optimization.

\section{\bettertogether: Alternating Weight and Prompt Optimization Steps for LM Programs}
\label{sec:methods}

We now introduce the \bettertogether\ algorithm, which alternates the weight and prompt optimization steps for LM programs.
We hypothesize that, when a large LM is used to teach itself how to tackle the task defined by an LM program, fine-tuning LM weights and prompts are \textit{both} essential to achieve the highest quality.
In particular, we expect that (1) prompt optimization before fine-tuning can lead to more successful datapoints for fine-tuning, and, (2) prompt optimization after fine-tuning can make adjustments to the behavior of the LM program, leading to higher quality outputs.
Considering that fine-tuning is often perceived as a more powerful tool, this can be surprising, especially when both approaches are ultimately applied over the same set of training inputs $X$.

\input{assets/algorithm_main.tex}

Accordingly, the general optimization framework for our algorithm is defined in \autoref{alg:main}.
Given a program $\Phi$, the algorithm begins by optimizing $\Phi$'s prompts, then fine-tuning its set of LM weights with the data bootstrapped using the optimized prompts, and finally optimizing its prompts again using the fine-tuned weights.
In principle, each of these steps could be treated as optional.
This will define the different possible combinations of \bettertogether\ that we will seek to evaluate in \autoref{sec:experiments}.
Specifically, we are interested in the quality of (1) the vanilla program $\Phi$ with simple user-supplied instructions as the prompts and no fine-tuning of $\mathbf{LM}$, (2) optimizing the prompts only, (3) optimizing the weights only, (4) optimizing the prompts twice, i.e., using the prompt-optimized $\Phi$ as a starting point for a second round of prompt optimization, (5) optimizing the weights twice, (6) optimizing the prompts then the weights, (7) vice versa, and (8) optimizing the prompts, weights, then prompts.
Overall, we expect the final three to consistently outperform the first five.

For \autoref{alg:main} to be complete, we need to instantiate Lines 1--3 with specific approaches for prompt optimization and LM fine-tuning. For this, we choose the \texttt{Bootstrap-$*$} family of algorithms from \citet{khattab2024dspy}, which work by executing an initial version of the program on input examples $(x_i, m_i) \in X$ and recording the inputs/outputs observed at each module when the \textit{final output} is ``correct'', i.e., $\mu(\Phi(x_i), m_i) \geq \lambda$ for some threshold $\lambda$ (e.g., $1.0$ for binary accuracy).
This is important to note: in line with our problem formulation, our prompt and weight optimization regimes are not simply training on hand-labeled data but on self-generated program traces.

\input{assets/algorithm_helper}

Instantiations for Lines~\mbox{1--3} of \autoref{alg:main} are shown in \autoref{alg:helper}.
For prompt optimization, we use \texttt{BootstrapFewshotRS} (\texttt{BFRS}) of DSPy, which self-generates potential few-shot examples of every module and applies a form of random search (\texttt{RS}) to select the specific generated few-shot examples that are used for prompting.
Overall, \texttt{BFRS} first divides $X$ into a training split $T$ and a validation split $V$ (Line 2). It then executes the provided $\LanguageProgram$ on the training inputs, collecting input--output pairs for every module in $\Phi$ for each $x_i \in T$.
This is called a trace $\tau$, and we keep only the traces assigned high scores by~$\mu$ (Line~4). Given all of these traces, \texttt{BFRS} samples multiple different subsets of a few traces $\tau'$ (Line~6), each of them containing a potential few-shot example for \textit{each} module in $\Phi$, and ultimately selects the subset that, when used to construct few-shot prompts (Line~7), achieves the highest score (Line~8).
This simple search strategy is known to consistently lead to large quality improvements in prompting LM programs~\cite{khattab2024dspy,opsahlong2024optimizing}, often outperforming manually or automatically optimizing prompt instructions or writing examples by hand.

For fine-tuning, we extend \texttt{BootstrapFinetune} (\texttt{BFT}) of DSPy, which, given a program $\Phi$, self-generates a large number examples for every module and combines them into one dataset to fine-tune the LM weights with an implicit multi-task objective, where the sub-tasks are the modules' roles.
Existing work has only considered \texttt{BFT} in a very narrow setting for LM programs: on \hotpotqa, \citet{khattab2024dspy} train a \texttt{T5-Large} model using traces from a few-shot \texttt{Llama2-13b} program, without considering getting an LM to teach itself via \texttt{BFT} nor considering a role for \texttt{BFRS} in the fine-tuned program.
In this work, we focus on allowing models to teach themselves and self-improve.
We propose for the first time combining the strategies of \texttt{BFRS} and \texttt{BFT} via alternation to get the same LM to teach itself better than either prompt or weight optimization in isolation.
One could test similar ideas in scenarios where a larger model does the bootstrapping for a smaller LM. This may lead to even better results but is outside our scope.

%% file: assets/algorithm_main.tex
\begin{algorithm}
    \caption{\bettertogether: Optimizing LM programs by alternating prompt and weight optimization steps, instantiated in \autoref{alg:helper}}
    \label{alg:main}
    \small
    
    \begin{algorithmic}
        \Require \textbf{Program} $\LanguageProgram = \LanguageProgramWeights \odot \LanguageProgramPrompts$, \\
        \hspace{8mm} with module weights $\Theta = [\theta_{1}, \ldots, \theta_{|\Phi|}]$ \\
        \hspace{8mm} and module prompts $\Pi = [\pi_{1}, \ldots, \pi_{|\Phi|}] $\\
        \textbf{Training Set} $X$ and \textbf{Metric} $\mu$
    \end{algorithmic}
    
    \begin{algorithmic}[1] %
        \Function{BetterTogether}{$\LanguageProgram$, $X$, $\mu$}
            \State $\Pi' \gets \textproc{OptimizePrompts($\Phi_{\langle \Theta, \Pi \rangle}$, $X$, $\mu$)}$
            \State $\Theta' \gets \textproc{FinetuneWeights($\Phi_{\langle \Theta, \Pi' \rangle}$, $X$, $\mu$)}$
            \State $\Pi'' \gets \textproc{OptimizePrompts($\Phi_{\langle \Theta', \Pi \rangle}$, $X$, $\mu$)}$
            \State \Return $\Phi_{\langle \Theta', \Pi'' \rangle}$
        \EndFunction
    \end{algorithmic}
\end{algorithm}

%% file: assets/algorithm_helper.tex
\begin{algorithm}
    \caption{Instantiating \autoref{alg:main}'s prompt \& weight optimizers with bootstrapping algorithms}\label{alg:helper}
    \small

    \begin{algorithmic}
        \Require \textbf{Training Set} $X$ and \textbf{Metric} $\mu$
    \end{algorithmic}
    
    \begin{algorithmic}[1] %
        \Function{BootstrapFewShotRS}{$\LanguageProgram$, $X$, $\mu$}
            \State $T, V \gets \textproc{SplitIntoTrainAndValidation}(X)$
            \State $\tau \gets \textproc{BootstrapTraces($\LanguageProgram$, $T$)}$
            \State $\tau \gets \textproc{FilterTraces($\tau$, $\mu$)}$ 
            \State Initialize attempts list $\mathcal{A} \gets \{\}$ 
            \For{$\tau' \in \textproc{SampleFewShotSubsets}(\tau)$}
                \State $\Pi' \gets \textproc{ConstructFewShotPrompts}(\tau')$
                \State $\sigma \gets \frac{1}{|V|} \sum_{\langle x_i, m_i \rangle \in V} \mu(\Phi_{\langle \Theta, \Pi' \rangle}(x_i), m_i)$
            \State Extend $\mathcal{A}$ with $(\sigma, \Pi')$
            \EndFor
            \State \Return $\Pi_{\max}$, $\mathcal{A}$'s highest-scoring prompts sequence
        \EndFunction
        \\

        \Function{BootstrapFinetune}{$\LanguageProgram$, $X$, $\mu$}
            \State $\tau \gets \textproc{BootstrapTraces($\LanguageProgram$, $X$)}$
            \State $\tau \gets \textproc{FilterTraces($\tau$, $\mu$)}$
            \State $\Theta' \gets \textproc{TrainLM($\tau$)}$
            \State \Return $\Theta'$
        \EndFunction
        \\

        \State Set \textproc{OptimizePrompts} as \textproc{BootstrapFewShotRS}
        \State Set \textproc{FinetuneWeights} as \textproc{BootstrapFinetune}
    \end{algorithmic}
\end{algorithm}

%% file: sections/03_experiments.tex
\section{Experimental Evaluation}
\label{sec:experiments}

\input{assets/table_main}

We now seek to evaluate our hypothesis on the importance of optimizing both LM weights and prompts of LM programs.
We conduct our evaluation across three datasets that span different tasks (and thus LM programs) each.
In particular, we use \hotpotqa~\cite{yang2018hotpotqa} for multi-hop reasoning, \gsmk~\cite{cobbe2021training} for arithmetic reasoning, and \iris~\cite{iris1988fisher} for classification.
We run our experiments using three models, \mistralshort~\cite{jiang2023mistral}, \llamatwoshort~\cite{touvron2023llama2}, \llamathreeshort~\cite{meta2024llama3}, keeping the model used for prompt optimization, bootstrapping training traces, and fine-tuning the same in a given experiment for all modules.
We initialize all the modules of an input program $\Phi$ to use the same LM weights (e.g. \mistralshort), but distinct prompt templates specialized for their particular module-level task, such as generating a search query or answering a question.
We implement all of our programs and optimizers as extensions to the DSPy framework.

For each dataset, we split the data into training, development, and test sets.
We shuffle the training set every time before we perform a prompt or weight optimization, controlled by random seed.
For prompt optimization, we sub-sample non-overlapping sets of training and validation examples from the initial training set for each task and use the \texttt{BFRS} prompt optimizer to optimize the module level prompts of a given $\Phi$, leaving its underlying LM weights untouched.
For weight optimization, we use all the available training examples as potential candidates to generate the data for fine-tuning $\Phi$'s LM weights and pass them to the \texttt{BFT} weight optimizer, which: (1) runs a given $\Phi$ on all the training examples, (2) keeps the traces where the final output of $\Phi$ was correct and filters out the rest, (3) gets module level prompt-completion pairs for each trace, (4) creates new pairs by replacing the prompt generated by $\Phi$ with a vanilla prompt, (5) combines all the module level prompt-completion pairs into one dataset, (6) fine-tunes $\Phi$'s LM weights on this dataset, and finally, (7) returns an updated $\Phi$ where the LM weights of all the modules are set to the fine-tuned LM.
We use the Low Rank Adaptation (LoRA; \citealt{hu2022lora}) method to fine-tune our LMs.

The full text for our programs and the vanilla prompts are shared in \autoref{app:programs}.
The license information for all LMs and datasets used as well as our implementation details such as hyperparameters and software are reported in \autoref{app:assets} and \autoref{app:implementation}, respectively.

\paragraph{Multi-hop Reasoning}
\hotpotqa{} (in the ``fullwiki'' setting) is a question answering task in which systems must find two Wikipedia page abstracts out of a corpus of 5 million via search and use them to answer a factoid question.
Therefore it can be implemented as a program that has three LM modules: the first two for generating \textit{search queries} (i.e., \textit{hops}) and the last one for generating an \textit{answer}.
Each module uses Chain-of-Thought (CoT; \citealt{wei2022chain}) to generate its outputs, producing a \textit{reasoning} string before the search query or the answer. 
Search queries are passed to a frozen \colberttwo~\cite{santhanam2022colbertv2} retriever.
Accuracy is measured using the exact match score of the answer with the ground truth answer for the given question, after normalizing case, stripping surrounding whitespace characters, and removing punctuation.
We use the following splits for \hotpotqa: \defaulttrain\ training examples and \defaultdev\ development examples drawn from the original training set, along with \hotpotqatest\ test examples drawn from the original validation set, since the original test set is not public.
We sub-sample \defaultprompttrain\ and  \defaultpromptval\ non-overlapping examples for the prompt optimization training and validation sets, respectively.

\paragraph{Arithmetic Reasoning}
\gsmk\ is a benchmark consisting of grade school math problems.
We implement it as an LM program with a single module using CoT prompting, where the LM generates a reasoning string followed by an answer.
We measure accuracy by extracting the last number from the first line of the model's response and comparing it to the ground truth response.
For \gsmk, we use a training set of \defaulttrain\ examples and a development set of \defaultdev\ examples, both sampled from the original training set.
We use all the \gsmktest\ test examples available for our test set.
We use \defaultprompttrain\ and  \defaultpromptval\ non-overlapping examples for training and validation sets used during prompt optimization.

\paragraph{Classification}
\iris\ is a classic classification task, where the goal is to classify species of \textit{Iris} flowers.
We use a single-module CoT DSPy program for \iris, with the goal of assessing whether it being a feature-based classification task gives a large advantage to methods based entirely on gradient descent (fine-tuning).
This tests the extrapolation of our hypothesis to a different setting from the other two tasks.
We measure accuracy using the exact match score of the answer and the correct answer given a question after normalizing both, as is the case for \hotpotqa.
The \iris\ dataset has a total of \iristotal\ examples, from which we create training, development, and test splits of 
equal size.
We sub-sample \iriscustomprompttrain\ and \iriscustompromptval\ non-overlapping examples to be used as the training and validation sets for prompt optimization, respectively.

%% file: assets/table_main.tex
\begin{table*}[ht]
    \centering
    \resizebox{\textwidth}{!}{
        \begin{tabular}{lccccccccc}
            \toprule
            \multirow{2}{*}{Strategy} & \multicolumn{3}{c}{\mistralshort} & \multicolumn{3}{c}{\llamatwoshort} & \multicolumn{3}{c}{\llamathreeshort} \\
            \cmidrule(lr){2-4} \cmidrule(lr){5-7} \cmidrule(lr){8-10}
            & \hotpotqa\ & \gsmk\ & \iris\ & \hotpotqa\ & \gsmk\ & \iris\ & \hotpotqa\ & \gsmk\ & \iris\ \\
            \midrule
            \textit{Baseline Strategies} \\
            \midrule
            Vanilla Zero-shot                         & 17.2           & 40.3           & 26.0           & 13.2           & 24.0           & \phantom{0}0.0 & 31.6           & 72.7           & 48.0           \\
            Prompt Optimization ($\Pi$)               & 33.8           & 46.4           & 57.3           & 33.3           & 26.0           & 56.7           & \textbf{46.9}  & 77.9           & 79.3           \\
            Weight Optimization ($\Theta$)            & 22.9           & 40.7           & 29.3           & 12.2           & 24.0           & --             & 34.8           & 75.1           & 37.3           \\
            $\Pi \rightarrow \Pi$                     & 33.8           & 47.7           & 59.3           & 32.6           & 24.7           & 64.0           & 46.5           & 77.6           & \textbf{82.0}  \\
            $\Theta \rightarrow \Theta$               & 24.0           & 42.8           & 38.0           & 13.0           & 24.1           & --             & 34.4           & 44.1           & 39.3           \\
            \midrule
            \textit{BetterTogether Strategies} \\
            \midrule
            $\Pi \rightarrow \Theta$                  & 36.3           & 47.3           & 30.7           & 32.7           & \textbf{27.3}  & 26.7           & 42.8           & 77.6          & 44.0            \\
            $\Theta \rightarrow \Pi$                  & 33.0           & \textbf{48.3}  & \textbf{66.7}  & 34.2           & 26.6           & --             & 43.6           & \textbf{78.9} & 78.7            \\
            $\Pi \rightarrow \Theta \rightarrow \Pi$  & \textbf{37.6}  & 46.8           & 52.7           & \textbf{34.8}  & 26.3           & \textbf{65.3}  & 46.7           & 77.0          & 79.3            \\
            \bottomrule
        \end{tabular}
    }
    \caption{
        \textbf{Main Results.} Percentage accuracies of baseline and \bettertogether\ strategies on \hotpotqa, \gsmk, and \iris\ evaluated on \mistralshort, \llamatwoshort, and \llamathreeshort.
        Reported are average performance of 3 runs on a held-out test set using different random seeds.
        \textbf{Bold} font is used to mark the highest score in a given column.
        Strategies where weight optimization is the first step use the vanilla (zero-shot) strategy to generate the initial fine-tuning dataset.
        If a model generates very few or no correct outputs under the vanilla strategy on the training set used to bootstrap the fine-tuning data, there will not be a sufficient dataset for fine-tuning.
        These settings are marked with ``--''.
    }
    \label{tab:main}
\end{table*}

%% file: sections/04_conclusion.tex
\section{Results \& Discussion}
\label{sec:conclusion}

Held-out test set performance of the strategies described in \autoref{sec:methods} is shared in \autoref{tab:main}.
Reported values are averaged across three runs with unique random seeds.
Results for the individual runs are reported separately in \autoref{app:extended_results}.

In \resultsuccessrate\ dataset and LM pairs, we observe that the best-performing strategies are the strategies that utilize prompt ($\Pi$) and weight ($\Theta$) optimization steps together, although there is no clear winner among the benchmarked \bettertogether\ strategies that optimize both.
Overall, optimizing prompts is essential on all the tasks, but optimizing prompts and weights together leads to strong gains over the best setting that only optimizes one of the two.

In summary, we have proposed to \textit{alternate between optimizing prompts and fine-tuning LM weights} and explored a few strategies for doing so. In experiments with multi-hop QA (\hotpotqa), mathematical reasoning (\gsmk), and feature-based classification (\iris), we show that our strategies are highly effective for getting an LM to teach itself to perform an LM program via bootstrapping, leading to \gainshotpotqaminmax\ gains for \hotpotqa, \gainsgsmkminmax\ gains for \gsmk, and \gainsirisminmax\ gains for \iris.

%% file: sections/05_limitations.tex
\section{Limitations}
\label{sec:limitations}

While this paper presents strong evidence from nine cases, spanning three tasks (and their corresponding LM programs) and three LMs, it is possible that other tasks, programs, or LMs will change the pattern in unforeseen ways.
In particular, we have only experimented with weight optimization in the form of LoRA fine-tuning of pre-trained models.
It is in principle possible that some other fine-tuning strategy would be so powerful and cost-effective as to remove the need for prompt optimization. 

In addition, though we expect our findings to inform many researchers and practitioners interested in optimizing LM programs, and encourage them to explore optimizing prompts and fine-tuning LM weights together, we do not yet understand \textit{why} both are important.
The role of prompt optimization and the role of fine-tuning in multi-stage LM programs are both new, and the relative lack of understanding of these roles in the emerging literature could pose risks in unanticipated interactions between these components, compared with standard gradient descent for neural networks, which has been studied for decades.

%% file: sections/__acknowledgments.tex
\section*{Acknowledgments}

D.S.\ is supported by Ravi Family Graduate Fellowship.
This work was partially supported by IBM as a founding member of the Stanford Institute for Human-Centered Artificial Intelligence (HAI), and by the HAI Hoffman--Yee Grant ``Dendritic Computation for Knowledge Systems''.

%% file: appendices/01_programs.tex
\section{Programs}
\label{app:programs}

The DSPy programs for \hotpotqa, \gsmk, and \iris\ are shared in Python Snippets~\ref{snippetpython:hotpotqa_program}, \ref{snippetpython:gsmk_program}, \ref{snippetpython:iris_program}, respectively.
Tables~\ref{tab:prompt_hotpotqa}, \ref{tab:prompt_gsmk}, \ref{tab:prompt_iris} show the vanilla (zero-shot) programs generated with these programs, along with sample model completions.

\input{assets/program_hotpotqa}
\input{assets/table_prompt_hotpotqa}

\input{assets/program_gsmk}
\input{assets/table_prompt_gsmk}

\input{assets/program_iris}
\input{assets/table_prompt_iris}

%% file: assets/program_hotpotqa.tex
\begin{snippetpython}[caption={DSPy program for \hotpotqa, consisting of three Chain-of-Thought (CoT) LM modules, along with a frozen retriever.
Each LM module in the program uses its own prompt and LM weights under the hood.}, label={snippetpython:hotpotqa_program}]
class HotPotQAProgram(dspy.Module):
    def __init__(self, passages_per_hop=3):
        super().__init__()

        self.retrieve = dspy.Retrieve(k=passages_per_hop)
        self.generate_query = [dspy.ChainOfThought("context, question -> search_query") for _ in range(2)]
        self.generate_answer = dspy.ChainOfThought("context, question -> answer")
    
    def forward(self, question):
        context = []
        
        for hop in range(2):
            search_query = self.generate_query[hop](context=context, question=question).search_query
            passages = self.retrieve(search_query).passages
            context = dsp.utils.deduplicate(context + passages)

        return self.generate_answer(context=context, question=question).copy(context=context)
\end{snippetpython}

%% file: assets/table_prompt_hotpotqa.tex
\begin{longtable}{r p{0.76\textwidth}}
    \caption{
        Vanilla (zero-shot) prompts for the \hotpotqa\ program given in \autoref{snippetpython:hotpotqa_program} with three LM modules, each with corresponding prompts and LM weights.
        \textbf{(1)} the \texttt{generate\_query[0]} module prompts the corresponding LM to generate a search query given a question along with an empty context; \textbf{(2)} the \texttt{generate\_query[1]} module prompts the corresponding model to generate a search query given the context paragraphs retrieved using the search query outputted by the prior module; \textbf{(3)} the \texttt{generate\_answer} module prompts the corresponding LM to answer the question given the context paragraphs retrieved by the earlier modules.
        The \textcolor{ModelCompletion}{green} text shows the model completions obtained by setting all the modules of the program to use \mistral\ as their corresponding LM.
        The correct answer to the question is indeed \texttt{Key deer} as predicted.
    } \\
    \toprule
    Module Label & Prompt \\
    \toprule
    \endfirsthead
    \toprule
    Module Label & Prompt \\
    \toprule
    \endhead
    \texttt{generate\_query[0]} & \input{assets/prompt_hotpotqa_module1} \\
    \midrule
    \texttt{generate\_query[1]} & \input{assets/prompt_hotpotqa_module2} \\
    \midrule
    \texttt{generate\_answer}   & \input{assets/prompt_hotpotqa_module3} \\
    \bottomrule
    \label{tab:prompt_hotpotqa}
\end{longtable}

%% file: assets/prompt_hotpotqa_module1.tex
\begin{unflavoredsnippet}
Given the fields `context`, `question`, produce the fields `search_query`.

---

Follow the following format.

Context: ${context}

Question: ${question}

Reasoning: Let's think step by step in order to ${produce the search_query}. We ...

Search Query: ${search_query}

---

Context: N/A

Question: What animal subject to the Foster's rule is the smallest North American deer?

Reasoning: Let's think step by step in order to #find the smallest North American deer subject to the Foster's rule.

1. The Foster's rule is a mathematical formula used to estimate the body weight of large mammals based on their ear surface area.
2. To apply the Foster's rule, we need to find the ear surface area of the deer species in question.
3. We don't have the ear surface area of each North American deer species in the context, so we need to search for it.
4. Therefore, our search query should include the keywords "North American deer," "ear surface area," and "Foster's rule."

Search Query: North American deer ear surface area Foster's rule smallest species#
\end{unflavoredsnippet}

%% file: assets/prompt_hotpotqa_module2.tex
\begin{unflavoredsnippet}
Given the fields `context`, `question`, produce the fields `search_query`.

---

Follow the following format.

Context: ${context}

Question: ${question}

Reasoning: Let's think step by step in order to ${produce the search_query}. We ...

Search Query: ${search_query}

---

Context:
[1] «Key deer | The Key deer ("Odocoileus virginianus clavium") is an endangered deer that lives only in the Florida Keys. It is a subspecies of the white-tailed deer ("O. virginianus"). It is the smallest North American deer.»
[2] «North American least shrew | The North American least shrew ("Cryptotis parva") is one of the smallest mammals, growing to be only up to 3 inches long. The North American least shrew has a long pointed snout and a tail never more than twice the length of its hind foot. It has a dense fur coat that is either grayish-brown or reddish-brown with a white belly. Its fur becomes lighter in the summer and darker in the winter. Although similar in appearance to several species of rodents, all shrews are members of the order Soricomorpha and should not be mistaken for a member of the Rodentia order. The North American least shrew's eyes are small and its ears are completely concealed within its short fur, giving it very poor eyesight and hearing.»
[3] «Wynnea americana | Wynnea americana, commonly known as moose antlers or rabbit ears, is a species of fungus in the family Sarcoscyphaceae. This uncommon inedible species is recognizable by its spoon-shaped or rabbit-ear shaped fruit bodies that may reach up to 13 cm tall. It has dark brown and warty outer surfaces, while the fertile spore-bearing inner surface is orange to pinkish to reddish brown. The fruit bodies grow clustered together from large underground masses of compacted mycelia known as sclerotia. In eastern North America, where it is typically found growing in the soil underneath hardwood trees, it is found from New York to Michigan south to Mexico. The species has also been collected from Costa Rica, India, and Japan.»

Question: What animal subject to the Foster's rule is the smallest North American deer?

Reasoning: Let's think step by step in order to #produce the search query. We need to find the smallest North American deer, which is mentioned in the context as the Key deer. Therefore, the search query should include the terms "Key deer" and "smallest North American deer".

Search Query: Key deer smallest North American deer#
\end{unflavoredsnippet}

%% file: assets/prompt_hotpotqa_module3.tex
\begin{unflavoredsnippet}
Given the fields `context`, `question`, produce the fields `answer`.

---

Follow the following format.

Context: ${context}

Question: ${question}

Reasoning: Let's think step by step in order to ${produce the answer}. We ...

Answer: ${answer}

---

Context:
[1] «Key deer | The Key deer ("Odocoileus virginianus clavium") is an endangered deer that lives only in the Florida Keys. It is a subspecies of the white-tailed deer ("O. virginianus"). It is the smallest North American deer.»
[2] «North American least shrew | The North American least shrew ("Cryptotis parva") is one of the smallest mammals, growing to be only up to 3 inches long. The North American least shrew has a long pointed snout and a tail never more than twice the length of its hind foot. It has a dense fur coat that is either grayish-brown or reddish-brown with a white belly. Its fur becomes lighter in the summer and darker in the winter. Although similar in appearance to several species of rodents, all shrews are members of the order Soricomorpha and should not be mistaken for a member of the Rodentia order. The North American least shrew's eyes are small and its ears are completely concealed within its short fur, giving it very poor eyesight and hearing.»
[3] «Wynnea americana | Wynnea americana, commonly known as moose antlers or rabbit ears, is a species of fungus in the family Sarcoscyphaceae. This uncommon inedible species is recognizable by its spoon-shaped or rabbit-ear shaped fruit bodies that may reach up to 13 cm tall. It has dark brown and warty outer surfaces, while the fertile spore-bearing inner surface is orange to pinkish to reddish brown. The fruit bodies grow clustered together from large underground masses of compacted mycelia known as sclerotia. In eastern North America, where it is typically found growing in the soil underneath hardwood trees, it is found from New York to Michigan south to Mexico. The species has also been collected from Costa Rica, India, and Japan.»
[4] «Pudú | The pudús (Mapudungun "püdü" or "püdu", Spanish: pudú , ] ) are two species of South American deer from the genus Pudu, and are the world's smallest deer. The name is a loanword from Mapudungun, the language of the indigenous Mapuche people of southern Chile and south-western Argentina. The two species of pudús are the northern pudú ("Pudu mephistophiles") from Venezuela, Colombia, Ecuador, and Peru, and the southern pudú ("Pudu puda"; sometimes incorrectly modified to "Pudu pudu") from southern Chile and south-western Argentina. Pudús range in size from 32 to tall, and up to 85 cm long. As of 2009, the southern pudu is classified as near threatened, while the northern pudu is classified as vulnerable in the IUCN Red List.»

Question: What animal subject to the Foster's rule is the smallest North American deer?

Reasoning: Let's think step by step in order to #find the answer. We know that the smallest North American deer is mentioned in the context. We also know that the Foster's rule refers to a group of animals. In this case, the Foster's rule refers to endangered deer species that live only in the Florida Keys. Therefore, the Key deer is the answer.

Answer: Key deer#
\end{unflavoredsnippet}

%% file: assets/program_gsmk.tex
\begin{snippetpython}[caption={DSPy program for \gsmk.}, label={snippetpython:gsmk_program}]
class CoTProgram(dspy.Module):
    def __init__(self):
        super().__init__()
        self.generate_answer = dspy.ChainOfThought("question -> answer")
    
    def forward(self, question):
        return self.generate_answer(question=question)
\end{snippetpython}

%% file: assets/table_prompt_gsmk.tex
\begin{longtable}{r p{0.76\textwidth}}
    \caption{
        Vanilla (zero-shot) prompts for the \gsmk\ program given in \autoref{snippetpython:gsmk_program} with one LM module.
        The \texttt{generate\_answer} module prompts the corresponding LM to answer a grade school math problem.
        The \textcolor{ModelCompletion}{green} text shows the model completion obtained by setting all the modules of the program to use \mistral\ as their corresponding LM.
        The model's response is evaluated as correct because the correct answer (\texttt{1200}) is contained in the first line of model's generation.
    } \\
    \toprule
    Module Label & Prompt \\
    \toprule
    \endfirsthead
    \toprule
    Module Label & Prompt \\
    \toprule
    \endhead
    \texttt{generate\_answer} & \input{assets/prompt_gsmk} \\
    \bottomrule
    \label{tab:prompt_gsmk}
\end{longtable}

%% file: assets/prompt_gsmk.tex
\begin{unflavoredsnippet}
Given the fields `question`, produce the fields `answer`.

---

Follow the following format.

Question: ${question}
Reasoning: Let's think step by step in order to ${produce the answer}. We ...
Answer: ${answer}

---

Question: Micah can type 20 words per minute and Isaiah can type 40 words per minute. How many more words can Isaiah type than Micah in an hour?
Reasoning: Let's think step by step in order to #find out how many more words Isaiah can type than Micah in an hour. We know that Micah can type 20 words per minute and Isaiah can type 40 words per minute. In one minute, Isaiah types twice as many words as Micah. In 60 minutes, Isaiah types 60 minutes * 40 words per minute = <<60*40=2400>>2400 words. Micah types 60 minutes * 20 words per minute = <<60*20=1200>>1200 words. The difference between the number of words Isaiah and Micah can type in an hour is 2400 words - 1200 words = <<2400-1200=1200>>1200 words.

Answer: Isaiah can type 1200 more words than Micah in an hour.#
\end{unflavoredsnippet}

%% file: assets/program_iris.tex
\begin{snippetpython}[caption={DSPy program for \iris, provided to us by the DSPy team.}, label={snippetpython:iris_program}]
class IrisSignature(dspy.Signature):
    "Given the petal and sepal dimensions in cm, predict the iris species."

    petal_length = dspy.InputField()
    petal_width = dspy.InputField()
    sepal_length = dspy.InputField()
    sepal_width = dspy.InputField()
    answer = dspy.OutputField(desc='setosa, versicolor, or virginica')

class IrisProgram(dspy.Module):
    def __init__(self):
        self.generate_answer = dspy.ChainOfThought(IrisSignature)
    
    def forward(self, petal_length, petal_width, sepal_length, sepal_width):
        return self.generate_answer(petal_length=petal_length, petal_width=petal_width, sepal_length=sepal_length, sepal_width=sepal_width)
\end{snippetpython}

%% file: assets/table_prompt_iris.tex
\begin{longtable}{r p{0.76\textwidth}}
    \caption{
        Vanilla (zero-shot) prompts for the \iris\ program given in \autoref{snippetpython:iris_program} with one LM module.
        The \texttt{generate\_answer} module prompts the corresponding LM to classify the \textit{Iris} flowers into one of the following species given their features: \textit{Setosa}, \textit{Versicolor}, \textit{Virginica}.
        The \textcolor{ModelCompletion}{green} text shows the model completion obtained by setting all the modules of the program to use \mistral\ as their corresponding LM.
        The correct answer is \texttt{Setosa}, just as predicted.
    }  \\
    \toprule
    Module Label & Prompt \\
    \toprule
    \endfirsthead
    \toprule
    Module Label & Prompt \\
    \toprule
    \endhead
    \texttt{generate\_answer} & \input{assets/prompt_iris} \\
    \bottomrule
    \label{tab:prompt_iris}
\end{longtable}

%% file: assets/prompt_iris.tex
\begin{unflavoredsnippet}
Given the petal and sepal dimensions in cm, predict the iris species.

---

Follow the following format.

Petal Length: ${petal_length}

Petal Width: ${petal_width}

Sepal Length: ${sepal_length}

Sepal Width: ${sepal_width}

Reasoning: Let's think step by step in order to ${produce the answer}. We ...

Answer: setosa, versicolor, or virginica

---

Petal Length: 1.4

Petal Width: 0.2

Sepal Length: 5.0

Sepal Width: 3.6

Reasoning: Let's think step by step in order to #predict the iris species. We will compare the given measurements with the average measurements of each iris species. 1. Setosa: The average petal length for setosa is 1.3 cm and the average petal width is 0.3 cm. The given petal length (1.4 cm) is slightly larger than the average, but the petal width (0.2 cm) is smaller than the average. However, the sepal dimensions (5.0 cm and 3.6 cm) are within the range of setosa. Based on these measurements, it is likely that the iris is setosa. 2. Versicolour: The average petal length for versicolour is 4.2 cm and the average petal width is 1.4 cm. The given petal length (1.4 cm) is smaller than the average, and the petal width (0.2 cm) is much smaller than the average. The sepal dimensions (5.0 cm and 3.6 cm) are also within the range of versicolour. However, the small petal dimensions suggest that it is less likely to be versicolour. 3. Virginica: The average petal length for virginica is 5.8 cm and the average petal width is 2.0 cm. The given petal length (1.4 cm) is much smaller than the average, and the petal width (0.2 cm) is much smaller than the average. The sepal dimensions (5.0 cm and 3.6 cm) are also within the range of virginica. However, the small petal dimensions suggest that it is less likely to be virginica. Based on the given measurements, it is most likely that the iris is setosa.

Answer: setosa.#
\end{unflavoredsnippet}

%% file: appendices/02_assets.tex
\section{Asset Information}
\label{app:assets}

We share the associated licenses for the models and datasets we used below.
For models, we list the specific \texttt{HuggingFace} model id we used to retrieve the respective weights.
\begin{enumerate}
    \item \mistral: \apachetwolicense
    \item \llamatwo: \llamatwocommunitylicense
    \item \llamathree: \llamathreecommunitylicense
    \item \hotpotqa: \apachetwolicense
    \item \gsmk: \mitlicense
    \item \iris: \ccbyfour
\end{enumerate}
All the LMs used in this work are intended for use in English.

%% file: appendices/03_implementation.tex
\section{Implementation Details}
\label{app:implementation}

In this section, we share the implementation details as it pertains to the LM inference and fine-tuning infrastructure, as well as compute requirements.

\paragraph{Inference} For inference, we host our models in \texttt{Docker}~\cite{merkel2014docker} instances through \texttt{HuggingFace}'s \texttt{text-generation-inference}~\cite{huggingface2023tgi} toolkit.
We keep the sampling parameters the same across all experiments, using \texttt{TopK} sampling with a \texttt{temperature} of \samplingtemperature, and  \texttt{top\_k}\ of \samplingtopk, until the model either generates a stopping string or a total of \samplingmaxtokens\ tokens including the tokens in the prompt.

\paragraph{Prompt Optimization} For prompt optimization, we use the \texttt{BootstrapFewShotRS} (\texttt{BFRS}) optimizer from the DSPy library.
In particular, we allow \texttt{BFRS} to randomly search \promptnumcandidateprograms\ candidate programs using up to \promptmaxlabeledemos\ few-shot examples for each module prompt.

\paragraph{Fine-tuning} For fine-tuning, we use Low Rank Adaptation (LoRA; \citealt{hu2022lora}) to train the \finetuneloratarget\ of our models, using a LoRA \texttt{rank} of \finetunelorar, \texttt{alpha} of \finetuneloraalpha, with no dropout.
We fine-tune all of our models for \finetunenumepochs\ epochs using \finetuneprecision\ precision, with a learning rate of \finetunelr\ and an \textit{effective batch size} of \finetuneeffectivebatchsize.{Compute}
The \bettertogether\ strategies explored in this paper are naturally more expensive to run when compared to just prompt optimizing or fine-tuning.
How these two steps compare to each other in terms of compute requirements or wall clock time depends on the particular settings used for each as well as the size of the dataset used.
Total approximate GPU hours to produce \autoref{tab:main} was $\approx$75 hours, using \gpuaonehundred\ GPUs.

%% file: appendices/04_extended_results.tex
\section{Extended Results}
\label{app:extended_results}

The results shared in \autoref{tab:main} are the average of three runs.
Tables \ref{tab:seeds_hotpotqa}, \ref{tab:seeds_gsmk}, and \ref{tab:seeds_iris} show the breakdown of the individual runs for \hotpotqa, \gsmk, and \iris, respectively.

\input{assets/table_seeds_hotpotqa}
\input{assets/table_seeds_gsmk}
\input{assets/table_seeds_iris}

%% file: assets/table_seeds_hotpotqa.tex
\begin{table}[ht]
    \centering
    \resizebox{\textwidth}{!}{
        \begin{tabular}{lcccccccccccc}
            \toprule
            \multirow{2}{*}{Strategy} & \multicolumn{4}{c}{\mistralshort} & \multicolumn{4}{c}{\llamatwoshort} & \multicolumn{4}{c}{\llamathreeshort} \\
            \cmidrule(lr){2-5} \cmidrule(lr){6-9} \cmidrule(lr){10-13}
            & Run 1 & Run 2 & Run 3 & Avg & Run 1 & Run 2 & Run 3 & Avg & Run 1 & Run 2 & Run 3 & Avg \\
            \midrule
            \textit{Baseline Strategies} \\
            \midrule
            Vanilla Zero-shot                         & 17.2 & 17.2 & 17.2 & 17.2          & 13.2 & 13.2 & 13.2 & 13.2          & 31.6 & 31.6 & 31.6 & 31.6          \\
            Prompt Optimization ($\Pi$)               & 32.7 & 34.7 & 34.0 & 33.8          & 33.3 & 33.3 & 33.4 & 33.3          & 45.7 & 47.4 & 47.5 & \textbf{46.9} \\
            Weight Optimization ($\Theta$)            & 22.0 & 23.1 & 23.5 & 22.9          & 12.4 & 11.8 & 12.3 & 12.2          & 34.9 & 35.3 & 34.3 & 34.8          \\
            $\Pi \rightarrow \Pi$                     & 31.7 & 36.0 & 33.7 & 33.8          & 31.7 & 33.1 & 33.1 & 32.6          & 47.3 & 45.4 & 46.7 & 46.5          \\
            $\Theta \rightarrow \Theta$               & 24.1 & 23.9 & 23.9 & 24.0          & 12.4 & 13.5 & 13.3 & 13.0          & 35.1 & 34.1 & 34.1 & 34.4          \\
            \midrule
            \textit{BetterTogether Strategies} \\
            \midrule
            $\Pi \rightarrow \Theta$                  & 34.9 & 39.1 & 34.9 & 36.3          & 32.8 & 32.3 & 33.1 & 32.7          & 40.6 & 42.1 & 45.7 & 42.8          \\
            $\Theta \rightarrow \Pi$                  & 29.3 & 33.8 & 35.8 & 33.0          & 36.0 & 33.4 & 33.1 & 34.2          & 44.5 & 40.9 & 45.3 & 43.6          \\
            $\Pi \rightarrow \Theta \rightarrow \Pi$  & 34.9 & 40.7 & 37.2 & \textbf{37.6} & 34.7 & 34.5 & 35.3 & \textbf{34.8} & 46.5 & 47.1 & 46.4 & 46.7          \\
            \bottomrule
        \end{tabular}
    }
    \caption{
        \textbf{Results of \hotpotqa\ Runs.}
        Percentage accuracies of baseline and \bettertogether\ strategies on \hotpotqa\ evaluated on \mistralshort, \llamatwoshort, and \llamathreeshort.
        Reported are performance of 3 runs on a held-out test set of \hotpotqatest\ examples, using different random seeds.
        \textbf{Bold} font in the average columns (Avg) is used to mark the highest score in a given column.
    }
    \label{tab:seeds_hotpotqa}
\end{table}

%% file: assets/table_seeds_gsmk.tex
\begin{table}[ht]
    \centering
    \resizebox{\textwidth}{!}{
        \begin{tabular}{lcccccccccccc}
            \toprule
            \multirow{2}{*}{Strategy} & \multicolumn{4}{c}{\mistralshort} & \multicolumn{4}{c}{\llamatwoshort} & \multicolumn{4}{c}{\llamathreeshort} \\
            \cmidrule(lr){2-5} \cmidrule(lr){6-9} \cmidrule(lr){10-13}
            & Run 1 & Run 2 & Run 3 & Avg & Run 1 & Run 2 & Run 3 & Avg & Run 1 & Run 2 & Run 3 & Avg \\
            \midrule
            \textit{Baseline Strategies} \\
            \midrule
            Vanilla Zero-shot                         & 40.3 & 40.3 & 40.3 & 40.3          & 24.0 & 24.0 & 24.0 & 24.0          & 72.7 & 72.7 & 72.7 & 72.7          \\
            Prompt Optimization ($\Pi$)               & 45.0 & 47.2 & 47.1 & 46.4          & 27.3 & 25.1 & 25.5 & 26.0          & 76.9 & 77.9 & 78.9 & 77.9          \\
            Weight Optimization ($\Theta$)            & 40.8 & 40.0 & 41.2 & 40.7          & 23.7 & 24.2 & 24.0 & 24.0          & 75.7 & 74.8 & 74.8 & 75.1          \\
            $\Pi \rightarrow \Pi$                     & 46.3 & 47.2 & 49.6 & 47.7          & 28.4 & 24.0 & 21.8 & 24.7          & 76.5 & 80.1 & 76.1 & 77.6          \\
            $\Theta \rightarrow \Theta$               & 42.9 & 41.8 & 43.8 & 42.8          & 24.0 & 24.3 & 24.0 & 24.1          & 52.2 & 36.6 & 43.4 & 44.0          \\
            \midrule
            \textit{BetterTogether Strategies} \\
            \midrule
            $\Pi \rightarrow \Theta$                  & 46.4 & 47.3 & 48.2 & 47.3          & 27.8 & 28.1 & 25.9 & \textbf{27.3} & 77.6 & 75.4 & 79.8 & 77.6          \\
            $\Theta \rightarrow \Pi$                  & 50.1 & 46.0 & 48.8 & \textbf{48.3} & 26.8 & 26.1 & 27.0 & 26.6          & 78.5 & 79.8 & 78.4 & \textbf{78.9} \\
            $\Pi \rightarrow \Theta \rightarrow \Pi$  & 44.9 & 48.5 & 47.1 & 46.8          & 27.1 & 25.9 & 25.9 & 26.3          & 77.6 & 75.4 & 77.8 & 77.0          \\
            \bottomrule
        \end{tabular}
    }
    \caption{
        \textbf{Results of \gsmk\ Runs.}
        Percentage accuracies of baseline and \bettertogether\ strategies on \gsmk\ evaluated on \mistralshort, \llamatwoshort, and \llamathreeshort.
        Reported are performance of 3 runs on a held-out test set of \gsmktest\ examples, using different random seeds.
        \textbf{Bold} font in the average columns (Avg) is used to mark the highest score in a given column.
    }
    \label{tab:seeds_gsmk}
\end{table}

%% file: assets/table_seeds_iris.tex
\begin{table}[ht]
    \centering
    \resizebox{\textwidth}{!}{
        \begin{tabular}{lcccccccccccc}
            \toprule
            \multirow{2}{*}{Strategy} & \multicolumn{4}{c}{\mistralshort} & \multicolumn{4}{c}{\llamatwoshort} & \multicolumn{4}{c}{\llamathreeshort} \\
            \cmidrule(lr){2-5} \cmidrule(lr){6-9} \cmidrule(lr){10-13}
            & Run 1 & Run 2 & Run 3 & Avg & Run 1 & Run 2 & Run 3 & Avg & Run 1 & Run 2 & Run 3 & Avg \\
            \midrule
            \textit{Baseline Strategies} \\
            \midrule
            Vanilla Zero-shot                         & 26.0 & 26.0 & 26.0 & 26.0           & \phantom{0}0.0 & \phantom{0}0.0 & \phantom{0}0.0 & \phantom{0}0.0 & 48.0 & 48.0 & 48.0 & 48.0           \\
            Prompt Optimization ($\Pi$)               & 52.0 & 54.0 & 66.0 & 57.3           & 44.0           & 68.0           & 58.0           & 56.7           & 62.0 & 96.0 & 80.0 & 79.3           \\
            Weight Optimization ($\Theta$)            & 24.0 & 34.0 & 30.0 & 29.3           & --             & --             & --             & --             & 38.0 & 40.0 & 34.0 & 37.3           \\
            $\Pi \rightarrow \Pi$                     & 48.0 & 64.0 & 66.0 & 59.3           & 66.0           & 70.0           & 56.0           & 64.0           & 70.0 & 94.0 & 82.0 & \textbf{82.0}  \\
            $\Theta \rightarrow \Theta$               & 40.0 & 36.0 & 38.0 & 38.0           & --             & --             & --             & --             & 44.0 & 36.0 & 38.0 & 39.3           \\
            \midrule
            \textit{BetterTogether Strategies} \\
            \midrule
            $\Pi \rightarrow \Theta$                  & 32.0 & 26.0 & 34.0 & 30.7           & 30.0           & 26.0           & 24.0           & 26.7           & 50.0 & 42.0 & 40.0 & 44.0           \\
            $\Theta \rightarrow \Pi$                  & 80.0 & 54.0 & 66.0 & \textbf{66.7}  & --             & --             & --             & --             & 78.0 & 78.0 & 80.0 & 78.7           \\
            $\Pi \rightarrow \Theta \rightarrow \Pi$  & 52.0 & 44.0 & 62.0 & 52.7           & 62.0           & 70.0           & 64.0           & \textbf{65.3}  & 74.0 & 80.0 & 84.0 & 79.3           \\
            \bottomrule
        \end{tabular}
    }
    \caption{
        \textbf{Results of \iris\ Runs.}
        Percentage accuracies of baseline and \bettertogether\ strategies on \iris\ evaluated on \mistralshort, \llamatwoshort, and \llamathreeshort.
        Reported are performance of 3 runs on a held-out test set of \iriscustomsplit\ examples, using different random seeds.
        \textbf{Bold} font in the average columns (Avg) is used to mark the highest score in a given column.
        Strategies where weight optimization is the first step use the vanilla (zero-shot) strategy to generate the initial fine-tuning dataset.
        If a model generates very few or no correct outputs under the vanilla strategy on the training set used to bootstrap the fine-tuning data, there will not be a sufficient dataset for fine-tuning.
        These settings are marked with ``--''.
    }
    \label{tab:seeds_iris}
\end{table}

%% file: main.bbl
\begin{thebibliography}{28}
\providecommand{\natexlab}[1]{#1}

\bibitem[{Beurer-Kellner et~al.(2023)Beurer-Kellner, Fischer, and Vechev}]{beurerkellner2023prompting}
Luca Beurer-Kellner, Marc Fischer, and Martin Vechev. 2023.
\newblock \href {https://doi.org/10.1145/3591300} {Prompting is programming: A query language for large language models}.
\newblock \emph{Proc. ACM Program. Lang.}, 7(PLDI).

\bibitem[{Cobbe et~al.(2021)Cobbe, Kosaraju, Bavarian, Chen, Jun, Kaiser, Plappert, Tworek, Hilton, Nakano, Hesse, and Schulman}]{cobbe2021training}
Karl Cobbe, Vineet Kosaraju, Mohammad Bavarian, Mark Chen, Heewoo Jun, Lukasz Kaiser, Matthias Plappert, Jerry Tworek, Jacob Hilton, Reiichiro Nakano, Christopher Hesse, and John Schulman. 2021.
\newblock \href {https://arxiv.org/abs/2110.14168} {Training verifiers to solve math word problems}.
\newblock \emph{Preprint}, arXiv:2110.14168.

\bibitem[{Dohan et~al.(2022)Dohan, Xu, Lewkowycz, Austin, Bieber, Lopes, Wu, Michalewski, Saurous, Sohl-dickstein, Murphy, and Sutton}]{dohan2022language}
David Dohan, Winnie Xu, Aitor Lewkowycz, Jacob Austin, David Bieber, Raphael~Gontijo Lopes, Yuhuai Wu, Henryk Michalewski, Rif~A. Saurous, Jascha Sohl-dickstein, Kevin Murphy, and Charles Sutton. 2022.
\newblock \href {https://arxiv.org/abs/2207.10342} {Language model cascades}.
\newblock \emph{Preprint}, arXiv:2207.10342.

\bibitem[{D'Oosterlinck et~al.(2024)D'Oosterlinck, Khattab, Remy, Demeester, Develder, and Potts}]{doosterlinck2024incontext}
Karel D'Oosterlinck, Omar Khattab, François Remy, Thomas Demeester, Chris Develder, and Christopher Potts. 2024.
\newblock \href {https://arxiv.org/abs/2401.12178} {In-context learning for extreme multi-label classification}.
\newblock \emph{Preprint}, arXiv:2401.12178.

\bibitem[{Fisher(1988)}]{iris1988fisher}
Ronald~A. Fisher. 1988.
\newblock \href {https://doi.org/10.24432/C56C76} {Iris}.
\newblock UCI Machine Learning Repository.

\bibitem[{Guu et~al.(2020)Guu, Lee, Tung, Pasupat, and Chang}]{guu2020realm}
Kelvin Guu, Kenton Lee, Zora Tung, Panupong Pasupat, and Ming{-}Wei Chang. 2020.
\newblock \href {http://proceedings.mlr.press/v119/guu20a.html} {Retrieval augmented language model pre-training}.
\newblock In \emph{Proceedings of the 37th International Conference on Machine Learning, {ICML} 2020, 13-18 July 2020, Virtual Event}, volume 119 of \emph{Proceedings of Machine Learning Research}, pages 3929--3938. {PMLR}.

\bibitem[{Hu et~al.(2022)Hu, Shen, Wallis, Allen{-}Zhu, Li, Wang, Wang, and Chen}]{hu2022lora}
Edward~J. Hu, Yelong Shen, Phillip Wallis, Zeyuan Allen{-}Zhu, Yuanzhi Li, Shean Wang, Lu~Wang, and Weizhu Chen. 2022.
\newblock \href {https://openreview.net/forum?id=nZeVKeeFYf9} {Lora: Low-rank adaptation of large language models}.
\newblock In \emph{The Tenth International Conference on Learning Representations, {ICLR} 2022, Virtual Event, April 25-29, 2022}. OpenReview.net.

\bibitem[{HuggingFace(2023)}]{huggingface2023tgi}
HuggingFace. 2023.
\newblock \href {https://huggingface.co/docs/text-generation-inference} {Text generation inference}.

\bibitem[{Jiang et~al.(2023{\natexlab{a}})Jiang, Sablayrolles, Mensch, Bamford, Chaplot, de~las Casas, Bressand, Lengyel, Lample, Saulnier, Lavaud, Lachaux, Stock, Scao, Lavril, Wang, Lacroix, and Sayed}]{jiang2023mistral}
Albert~Q. Jiang, Alexandre Sablayrolles, Arthur Mensch, Chris Bamford, Devendra~Singh Chaplot, Diego de~las Casas, Florian Bressand, Gianna Lengyel, Guillaume Lample, Lucile Saulnier, Lélio~Renard Lavaud, Marie-Anne Lachaux, Pierre Stock, Teven~Le Scao, Thibaut Lavril, Thomas Wang, Timothée Lacroix, and William~El Sayed. 2023{\natexlab{a}}.
\newblock \href {https://arxiv.org/abs/2310.06825} {Mistral 7b}.
\newblock \emph{Preprint}, arXiv:2310.06825.

\bibitem[{Jiang et~al.(2023{\natexlab{b}})Jiang, Xu, Gao, Sun, Liu, Dwivedi-Yu, Yang, Callan, and Neubig}]{jiang-etal-2023-active}
Zhengbao Jiang, Frank Xu, Luyu Gao, Zhiqing Sun, Qian Liu, Jane Dwivedi-Yu, Yiming Yang, Jamie Callan, and Graham Neubig. 2023{\natexlab{b}}.
\newblock \href {https://doi.org/10.18653/v1/2023.emnlp-main.495} {Active retrieval augmented generation}.
\newblock In \emph{Proceedings of the 2023 Conference on Empirical Methods in Natural Language Processing}, pages 7969--7992, Singapore. Association for Computational Linguistics.

\bibitem[{Khattab et~al.(2021)Khattab, Potts, and Zaharia}]{khattab2021baleen}
Omar Khattab, Christopher Potts, and Matei~A. Zaharia. 2021.
\newblock \href {https://proceedings.neurips.cc/paper/2021/hash/e8b1cbd05f6e6a358a81dee52493dd06-Abstract.html} {Baleen: Robust multi-hop reasoning at scale via condensed retrieval}.
\newblock In \emph{Advances in Neural Information Processing Systems 34: Annual Conference on Neural Information Processing Systems 2021, NeurIPS 2021, December 6-14, 2021, virtual}, pages 27670--27682.

\bibitem[{Khattab et~al.(2022)Khattab, Santhanam, Li, Hall, Liang, Potts, and Zaharia}]{khattab2022demonstrate}
Omar Khattab, Keshav Santhanam, Xiang~Lisa Li, David Hall, Percy Liang, Christopher Potts, and Matei Zaharia. 2022.
\newblock \href {https://arxiv.org/abs/2212.14024} {Demonstrate-search-predict: Composing retrieval and language models for knowledge-intensive nlp}.
\newblock \emph{ArXiv preprint}, abs/2212.14024.

\bibitem[{Khattab et~al.(2024)Khattab, Singhvi, Maheshwari, Zhang, Santhanam, A, Haq, Sharma, Joshi, Moazam, Miller, Zaharia, and Potts}]{khattab2024dspy}
Omar Khattab, Arnav Singhvi, Paridhi Maheshwari, Zhiyuan Zhang, Keshav Santhanam, Sri~Vardhamanan A, Saiful Haq, Ashutosh Sharma, Thomas~T. Joshi, Hanna Moazam, Heather Miller, Matei Zaharia, and Christopher Potts. 2024.
\newblock \href {https://openreview.net/forum?id=sY5N0zY5Od} {{DSP}y: Compiling declarative language model calls into state-of-the-art pipelines}.
\newblock In \emph{The Twelfth International Conference on Learning Representations}.

\bibitem[{Lewis et~al.(2020)Lewis, Perez, Piktus, Petroni, Karpukhin, Goyal, K{\"{u}}ttler, Lewis, Yih, Rockt{\"{a}}schel, Riedel, and Kiela}]{lewis2020retrieval}
Patrick S.~H. Lewis, Ethan Perez, Aleksandra Piktus, Fabio Petroni, Vladimir Karpukhin, Naman Goyal, Heinrich K{\"{u}}ttler, Mike Lewis, Wen{-}tau Yih, Tim Rockt{\"{a}}schel, Sebastian Riedel, and Douwe Kiela. 2020.
\newblock \href {https://proceedings.neurips.cc/paper/2020/hash/6b493230205f780e1bc26945df7481e5-Abstract.html} {Retrieval-augmented generation for knowledge-intensive {NLP} tasks}.
\newblock In \emph{Advances in Neural Information Processing Systems 33: Annual Conference on Neural Information Processing Systems 2020, NeurIPS 2020, December 6-12, 2020, virtual}.

\bibitem[{Ma et~al.(2023)Ma, Gong, He, Zhao, and Duan}]{ma-etal-2023-query}
Xinbei Ma, Yeyun Gong, Pengcheng He, Hai Zhao, and Nan Duan. 2023.
\newblock \href {https://doi.org/10.18653/v1/2023.emnlp-main.322} {Query rewriting in retrieval-augmented large language models}.
\newblock In \emph{Proceedings of the 2023 Conference on Empirical Methods in Natural Language Processing}, pages 5303--5315, Singapore. Association for Computational Linguistics.

\bibitem[{Merkel(2014)}]{merkel2014docker}
Dirk Merkel. 2014.
\newblock Docker: lightweight linux containers for consistent development and deployment.
\newblock \emph{Linux J.}, 2014(239).

\bibitem[{MetaAI(2024)}]{meta2024llama3}
MetaAI. 2024.
\newblock \href {https://ai.meta.com/blog/meta-llama-3/} {Meta llama 3}.

\bibitem[{Opsahl-Ong et~al.(2024)Opsahl-Ong, Ryan, Purtell, Broman, Potts, Zaharia, and Khattab}]{opsahlong2024optimizing}
Krista Opsahl-Ong, Michael~J Ryan, Josh Purtell, David Broman, Christopher Potts, Matei Zaharia, and Omar Khattab. 2024.
\newblock \href {https://arxiv.org/abs/2406.11695} {Optimizing instructions and demonstrations for multi-stage language model programs}.

\bibitem[{Pourreza and Rafiei(2023)}]{pourreza2023dinsql}
Mohammadreza Pourreza and Davood Rafiei. 2023.
\newblock \href {http://papers.nips.cc/paper\_files/paper/2023/hash/72223cc66f63ca1aa59edaec1b3670e6-Abstract-Conference.html} {{DIN-SQL:} decomposed in-context learning of text-to-sql with self-correction}.
\newblock In \emph{Advances in Neural Information Processing Systems 36: Annual Conference on Neural Information Processing Systems 2023, NeurIPS 2023, New Orleans, LA, USA, December 10 - 16, 2023}.

\bibitem[{Press et~al.(2023)Press, Zhang, Min, Schmidt, Smith, and Lewis}]{press2023measuring}
Ofir Press, Muru Zhang, Sewon Min, Ludwig Schmidt, Noah Smith, and Mike Lewis. 2023.
\newblock \href {https://doi.org/10.18653/v1/2023.findings-emnlp.378} {Measuring and narrowing the compositionality gap in language models}.
\newblock In \emph{Findings of the Association for Computational Linguistics: EMNLP 2023}, pages 5687--5711, Singapore. Association for Computational Linguistics.

\bibitem[{Qi et~al.(2021)Qi, Lee, Sido, and Manning}]{qi2021retrieve}
Peng Qi, Haejun Lee, Tg~Sido, and Christopher Manning. 2021.
\newblock \href {https://doi.org/10.18653/v1/2021.emnlp-main.292} {Answering open-domain questions of varying reasoning steps from text}.
\newblock In \emph{Proceedings of the 2021 Conference on Empirical Methods in Natural Language Processing}, pages 3599--3614, Online and Punta Cana, Dominican Republic. Association for Computational Linguistics.

\bibitem[{Santhanam et~al.(2022)Santhanam, Khattab, Saad-Falcon, Potts, and Zaharia}]{santhanam2022colbertv2}
Keshav Santhanam, Omar Khattab, Jon Saad-Falcon, Christopher Potts, and Matei Zaharia. 2022.
\newblock \href {https://doi.org/10.18653/v1/2022.naacl-main.272} {{C}ol{BERT}v2: Effective and efficient retrieval via lightweight late interaction}.
\newblock In \emph{Proceedings of the 2022 Conference of the North American Chapter of the Association for Computational Linguistics: Human Language Technologies}, pages 3715--3734, Seattle, United States. Association for Computational Linguistics.

\bibitem[{Schlag et~al.(2023)Schlag, Sukhbaatar, Celikyilmaz, tau Yih, Weston, Schmidhuber, and Li}]{schlag2023large}
Imanol Schlag, Sainbayar Sukhbaatar, Asli Celikyilmaz, Wen tau Yih, Jason Weston, Jürgen Schmidhuber, and Xian Li. 2023.
\newblock \href {https://arxiv.org/abs/2305.05364} {Large language model programs}.
\newblock \emph{Preprint}, arXiv:2305.05364.

\bibitem[{Touvron et~al.(2023)Touvron, Martin, Stone, Albert, Almahairi, Babaei, Bashlykov, Batra, Bhargava, Bhosale, Bikel, Blecher, Ferrer, Chen, Cucurull, Esiobu, Fernandes, Fu, Fu, Fuller, Gao, Goswami, Goyal, Hartshorn, Hosseini, Hou, Inan, Kardas, Kerkez, Khabsa, Kloumann, Korenev, Koura, Lachaux, Lavril, Lee, Liskovich, Lu, Mao, Martinet, Mihaylov, Mishra, Molybog, Nie, Poulton, Reizenstein, Rungta, Saladi, Schelten, Silva, Smith, Subramanian, Tan, Tang, Taylor, Williams, Kuan, Xu, Yan, Zarov, Zhang, Fan, Kambadur, Narang, Rodriguez, Stojnic, Edunov, and Scialom}]{touvron2023llama2}
Hugo Touvron, Louis Martin, Kevin Stone, Peter Albert, Amjad Almahairi, Yasmine Babaei, Nikolay Bashlykov, Soumya Batra, Prajjwal Bhargava, Shruti Bhosale, Dan Bikel, Lukas Blecher, Cristian~Canton Ferrer, Moya Chen, Guillem Cucurull, David Esiobu, Jude Fernandes, Jeremy Fu, Wenyin Fu, Brian Fuller, Cynthia Gao, Vedanuj Goswami, Naman Goyal, Anthony Hartshorn, Saghar Hosseini, Rui Hou, Hakan Inan, Marcin Kardas, Viktor Kerkez, Madian Khabsa, Isabel Kloumann, Artem Korenev, Punit~Singh Koura, Marie-Anne Lachaux, Thibaut Lavril, Jenya Lee, Diana Liskovich, Yinghai Lu, Yuning Mao, Xavier Martinet, Todor Mihaylov, Pushkar Mishra, Igor Molybog, Yixin Nie, Andrew Poulton, Jeremy Reizenstein, Rashi Rungta, Kalyan Saladi, Alan Schelten, Ruan Silva, Eric~Michael Smith, Ranjan Subramanian, Xiaoqing~Ellen Tan, Binh Tang, Ross Taylor, Adina Williams, Jian~Xiang Kuan, Puxin Xu, Zheng Yan, Iliyan Zarov, Yuchen Zhang, Angela Fan, Melanie Kambadur, Sharan Narang, Aurelien Rodriguez, Robert Stojnic, Sergey Edunov, and Thomas
  Scialom. 2023.
\newblock \href {https://arxiv.org/abs/2307.09288} {Llama 2: Open foundation and fine-tuned chat models}.

\bibitem[{Viswanathan et~al.(2023)Viswanathan, Zhao, Bertsch, Wu, and Neubig}]{viswanathan2023prompt2model}
Vijay Viswanathan, Chenyang Zhao, Amanda Bertsch, Tongshuang Wu, and Graham Neubig. 2023.
\newblock \href {https://doi.org/10.18653/v1/2023.emnlp-demo.38} {{P}rompt2{M}odel: Generating deployable models from natural language instructions}.
\newblock In \emph{Proceedings of the 2023 Conference on Empirical Methods in Natural Language Processing: System Demonstrations}, pages 413--421, Singapore. Association for Computational Linguistics.

\bibitem[{Wei et~al.(2022)Wei, Wang, Schuurmans, Bosma, Ichter, Xia, Chi, Le, and Zhou}]{wei2022chain}
Jason Wei, Xuezhi Wang, Dale Schuurmans, Maarten Bosma, Brian Ichter, Fei Xia, Ed~H. Chi, Quoc~V. Le, and Denny Zhou. 2022.
\newblock \href {http://papers.nips.cc/paper\_files/paper/2022/hash/9d5609613524ecf4f15af0f7b31abca4-Abstract-Conference.html} {Chain-of-thought prompting elicits reasoning in large language models}.
\newblock In \emph{Advances in Neural Information Processing Systems 35: Annual Conference on Neural Information Processing Systems 2022, NeurIPS 2022, New Orleans, LA, USA, November 28 - December 9, 2022}.

\bibitem[{Yang et~al.(2018)Yang, Qi, Zhang, Bengio, Cohen, Salakhutdinov, and Manning}]{yang2018hotpotqa}
Zhilin Yang, Peng Qi, Saizheng Zhang, Yoshua Bengio, William Cohen, Ruslan Salakhutdinov, and Christopher~D. Manning. 2018.
\newblock \href {https://doi.org/10.18653/v1/D18-1259} {{H}otpot{QA}: A dataset for diverse, explainable multi-hop question answering}.
\newblock In \emph{Proceedings of the 2018 Conference on Empirical Methods in Natural Language Processing}, pages 2369--2380, Brussels, Belgium. Association for Computational Linguistics.

\bibitem[{Zelikman et~al.(2022)Zelikman, Wu, Mu, and Goodman}]{zelikman2022star}
Eric Zelikman, Yuhuai Wu, Jesse Mu, and Noah~D. Goodman. 2022.
\newblock \href {http://papers.nips.cc/paper\_files/paper/2022/hash/639a9a172c044fbb64175b5fad42e9a5-Abstract-Conference.html} {Star: Bootstrapping reasoning with reasoning}.
\newblock In \emph{Advances in Neural Information Processing Systems 35: Annual Conference on Neural Information Processing Systems 2022, NeurIPS 2022, New Orleans, LA, USA, November 28 - December 9, 2022}.

\end{thebibliography}
